\documentclass[runningheads]{llncs}
\usepackage{graphicx}
\usepackage{subcaption}
\usepackage{multirow}
\usepackage{amsmath} 
\usepackage{array}

\begin{document}
\title{Comparison of Machine Learning Methods for Assigning Software Issues to Team Members}
%
%
\author{Büşra Tabak\inst{1}\orcidID{0000-0001-7460-3689} \and \\
Fatma Başak Aydemir\inst{1}\orcidID{0000-0003-3833-3997} 
}

\authorrunning{Tabak and Aydemir}
%
\institute{Boğaziçi University, Turkey\\
\email\{busra.tabak, basak.aydemir\}@boun.edu.tr}
\maketitle              
\begin{abstract}
Software issues contain units of work to fix, improve, or create new threads during the development and facilitate communication among the team members. 
Assigning an issue to the most relevant team member and determining a category of an issue is a tedious and challenging task. Wrong classifications cause delays and rework in the project and trouble among the team members. 
This paper proposes a set of carefully curated linguistic features for shallow machine learning methods and compares the performance of shallow and ensemble methods with deep language models. Unlike the state-of-the-art, we assign issues to four roles (designer, developer, tester, and leader) rather than to specific individuals or teams to contribute to the generality of our solution. We also consider the level of experience of the developers to reflect the industrial practices in our solution formulation. We collect and annotate five industrial data sets from one of the top three global television producers to evaluate our proposal and compare it with deep language models. 
Our data sets contain 5324 issues in total. We show that an ensemble classifier of shallow techniques achieves 0.92 for issue assignment in accuracy which is statistically comparable to the state-of-the-art deep language models. 
The contributions include the public sharing of five annotated industrial issue data sets, the development of a clear and comprehensive feature set, the introduction of a novel label set, and the validation of the efficacy of an ensemble classifier of shallow machine learning techniques.

\end{abstract}

\section{Introduction}
\label{sec:intro}

Software project development refers to the process of creating a software product from start to finish, including planning, designing, coding, testing, and maintenance. It involves a team of developers, often with different specializations, working together to produce a working software product. Software project management involves overseeing the development process, ensuring that the project is completed on time, within budget, and to the expected quality standards \cite{Bertram2010}. This includes managing resources, schedules, and budgets, as well as communicating with stakeholders and ensuring that the project meets its objectives. Effective project management is necessary for successful software development.

One of the primary responsibilities of a project manager is to identify and address software issues as they arise throughout the development process \cite{Stellman2005applied}. These issues can include technical challenges, quality assurance problems, or unexpected delays. The project manager must work with the development team to find solutions to these issues, prioritize tasks, and make adjustments to the project plan as needed. By effectively managing software issues, the project manager can help ensure that the development process stays on track, that the software product is delivered on time and to the expected quality standards, and that the project stays within budget.

Issue Tracking Systems (ITS) are designed to help development teams track and manage software issues throughout the development process. These systems allow developers to identify, report, and prioritize software issues and assign them to team members for resolution \cite{Merten2016software}. Issue tracking systems often include features such as issue tracking, bug reporting, status tracking, and reporting tools, enabling developers to manage issues effectively and ensure that they are resolved in a timely manner. Issues can be created by users with different roles such as software developers, team leaders, testers, or even customer support teams in these tools. Bertram et al. \cite{Bertram2010} carry out a qualitative study of ITS as used by small software development teams. They demonstrate that ITS plays a key critical role in the communication and collaboration within software development teams based on their interviews with various stakeholders.

Text classification is an important problem that is the task of assigning a label to a given text \cite{Sebastiani2002}. Text classification has started to be used as a tool to produce solutions in many studies in various fields due to the abundance and diversity of data known as big data. The main focus of this paper is to address the issue classification problem through an issue assignment approach where we assign the identified issues to appropriate team members or departments for further resolution. To accomplish this, we treat the problem as a text classification challenge. We leverage machine learning algorithms and natural language processing techniques to analyze and classify the text data of the issues. By applying these techniques, we are able to extract relevant information from the issue descriptions, such as the issue severity, context, and other important details. Overall, by tackling the issue classification problem through this approach, we aim to provide a more comprehensive and effective solution for issue management and resolution.

The issue assignment approach enables us to allocate the issues to the most suitable team members or departments. This helps to streamline the resolution process and ensure that the issues are addressed by the right people, thereby improving the overall efficiency and effectiveness of the support system. We decide that assigning issues to groups of employees who can perform the same activities is preferable to the individuals. Some employees in the issue history may not have been able to complete the task that is automatically assigned to them in that planning time due to a variety of factors, including seasonal spikes in workload, illness, or employee turnover \cite{Jonsson2016}. To effectively manage the employees in our data set, we have grouped them based on the fields they work in. This approach has resulted in the identification of four main teams in the data set, namely the software developer, UI/UX designer, software tester, and team leader. The software developer team represents the majority of the data set, making them a crucial focus of our analysis. To improve time management and issue resolution, it is important to assign the right issues to the right developers. To achieve this, we have categorized the Software Developers using sub-labels that are generally accepted in the industry, such as senior, mid, and junior software developer levels. This categorization helps us identify the experience level and skill set of each developer, allowing us to allocate the most appropriate tasks to each team member. These teams may differ according to the project or the company. For example, new teams such as business analysts, and product owners can be added or some teams can be split or removed. At this point, we expect the side that will use the system to separate the individuals according to the teams. After a newly opened issue is automatically classified among these classes, it can be randomly assigned to the individuals in the relevant team, or the individuals in the team or the team leader can make this assignment manually. 

In our study, we use a closed-source data set for our analysis contrary to the majority of studies in the literature. We obtain five projects from the company's Jira interface for analysis. We focus exclusively on the main language of the issues. To prepare the data set for this study, we determine the label values by changing the people assigned to the issue according to the fields they work in, based on information we receive from the company.

ITS often contain a wealth of valuable data related to software issues. In our study, we set out to analyze this data using NLP methods, with the goal of creating a feature set that would be simpler and more successful than the word embedding methods typically used in text classification. To create our feature set, we use a range of NLP techniques to analyze the language used in software issues like part-of-speech tagging and sentiment analysis. We then compare our feature set with commonly used word embedding methods and apply a range of machine learning, ensemble learning, and deep-learning techniques to our annotated data set. This allows us to evaluate the efficiency of our approach using a range of standard metrics, including accuracy, precision, recall, and F1-score.

We have made several significant contributions to the state of the art in issue classification. 
\begin{itemize}
    \item Data set: We provide a closed-source issue data set from the industry in both Turkish and English. This data set is publicly available for further research, and to the best of our knowledge, there is no shared commercial issue data set for both languages in the literature.
    \item Feature set: We develop an understandable feature set that is extracted from the information in the issues, which can be applied to all issue classification procedures with high accuracy and low complexity. 
     \item Label set: We introduce novel labels for issue assignment. By incorporating these new labels, we expand the boundaries of current research and offer unique insights into the underlying themes, contributing to a more comprehensive understanding of the domain.
\end{itemize}

The remainder of this paper is structured as follows: 
 Section \ref{sec:background} describes the background of this study including the structure of software issues in issue tracking systems. In Section \ref{sec:approach}, we present our experimental setup and approach, followed by our results and analysis in Section \ref{sec:results}. In Section \ref{sec:discuss}, we discuss the threats to validity and user evaluation. In Section \ref{sec:relwork}, we discuss related work and similar classification endeavors with Turkish issue reports. Section \ref{sec:conclusions} concludes our work and discusses future work.

\section{Background}
\label{sec:background}
To understand our approach, it is essential to have a solid understanding of the structure of software issues in Issue Tracking Systems (ITS).

\subsection{Software Issues in ITS}
This section explores software issues within ITS, covering their anatomy, life cycle, and significance in software projects. 

\subsubsection{Issue Tracking Systems (ITS)}
An issue tracking system (ITS) is a software application that manages and maintains lists of issues. It provides a database of issue reports for a software project. Members of the software development team stay aligned and collaborate faster with the help of these tools. Users can log in with a password and post a new issue report or comment on an already-posted issue. There are various issue-tracking tools that are used in software development projects, such as JIRA, Bugzilla, and Azure DevOps. Although there are some differences among these tools, they all share a similar basic structure. We developed our approach with a data set of a company that uses JIRA software.

\subsubsection{Anatomy of an Issue}

\begin{figure}
\includegraphics[width=\textwidth]{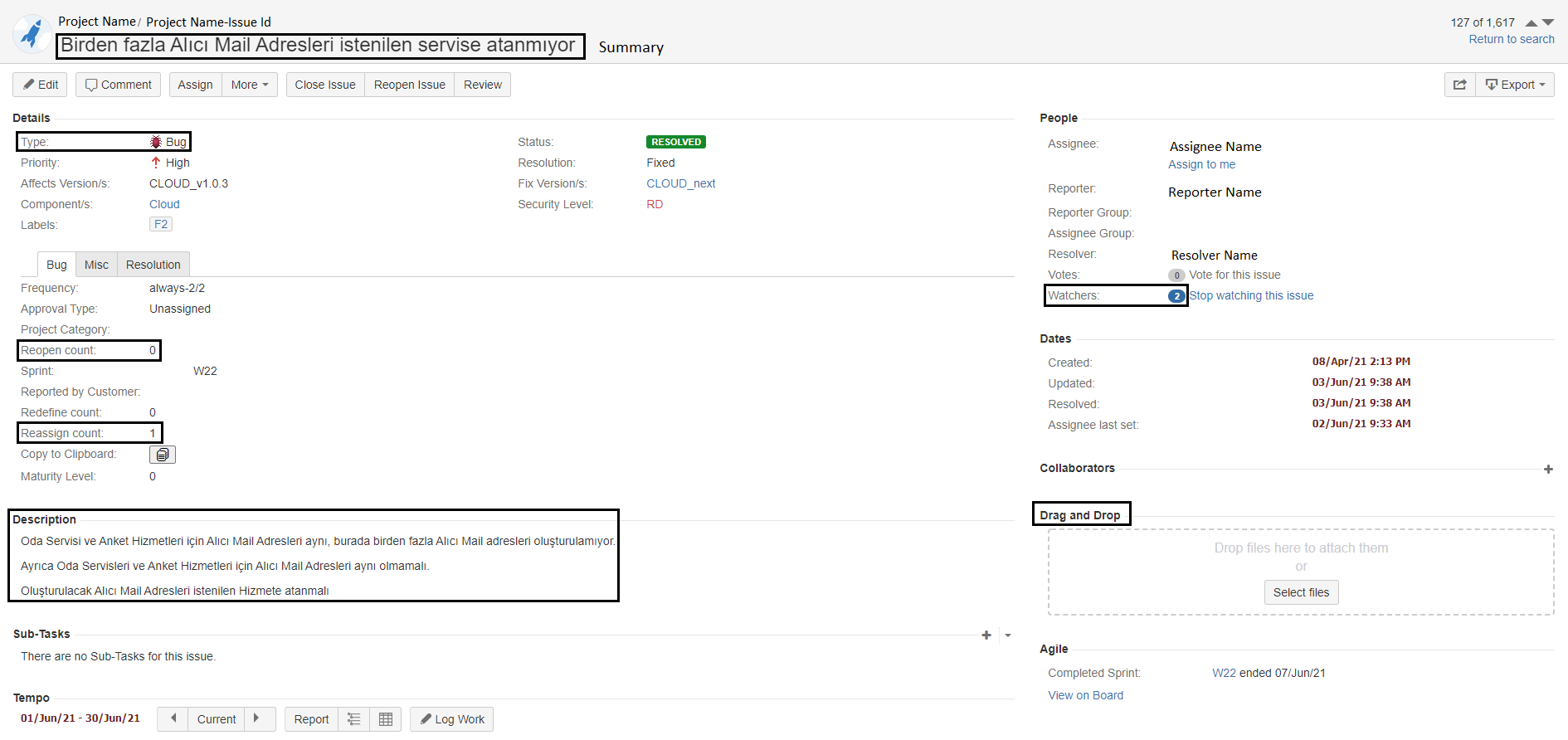}
\caption{An example issue from the Jira interface of the company.} \label{fig:issueSample}
\end{figure}

The issue report in ITS contains various fields, each of which contains a different piece of information as shown in Figure \ref{fig:issueSample}. The following details \cite{Singh2011} could typically be included in the generic process of reporting an issue:  
\newpage
\begin{itemize}
    \item Summary: Title of the issue.
    \item Description: Issue details including its cause, location, and timing.
    \item Version: Version of the project to which the issue belongs.
    \item Component: The predetermined component on which the issue depends.
    \item Attachment: Attaches such as pictures, videos, and documents uploaded to the issue.
    \item Priority: Urgency level of the issue. 
    \item Severity: Frequent occurrence and systemic effects.
    \item Status: Current status of the issue.
    \item Reporter: Person who creates the issue.
    \item Assignee: Person to whom the issue is assigned.
    \item Revision History: Historical changes of the issue.
    \item Estimated time: Estimated time spent to develop or solve the issue.
    \item Comments: Additional details that can be used to understand the issue.
\end{itemize}

The summary, description, and comments all contain textual details about the issue. In our methodology, we extract numerical features using the language structure of summary and description. Version is the current version of the project which differs with each release. The issue can have a predefined tag or component added to it. (e.g. project\_v1.1.0) Users can upload files to help others understand the issue, an attachment refers to it. Priority, severity, and status are the categorical features of the issue. Priority is the urgency level, severity is the 
frequency of occurrence and status is the current status of the issue such as committed or done. People play different roles as they interact with reports in ITS. The person who submits the report is the reporter and the assignee is the person to whom the issue is assigned. If the issue has been reported previously, historical changes are shown in the revision history. The estimated time is the time spent on the development and varies depending on the effort metric. The fields offered by each ITS vary, and not all fields are filled out for each project. Our strategy makes use of the fields that are largely filled in.

\subsubsection{Life Cycle of an Issue}
\label{sec:lifecycle_of_an_issue}

\begin{figure}
\includegraphics[width=\textwidth]{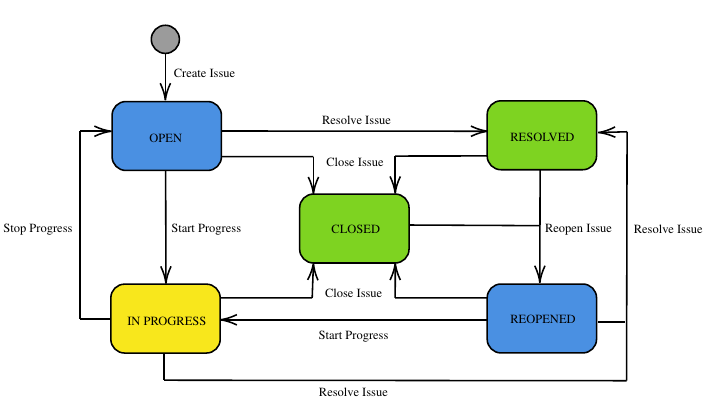}
\caption{Jira workflow.} \label{fig:workflow}
\end{figure}

The series of phases and phase changes that an issue experiences throughout its life cycle is referred to as a ``workflow." Workflows for issues typically represent development cycles and business processes. Figure \ref{fig:workflow} shows a standard workflow of JIRA. The following stages of the JIRA workflow must be monitored as soon as an issue is created:
\begin{itemize}
    \item Open: The issue is open after creation and can be assigned to the assignee to begin working on it. 
    \item In Progress: The assignee has taken the initiative to begin working on the issue. 
    \item Resolved: The issue's sub-tasks and works have all been finished. The reporter is currently waiting to confirm the matter. If verification is successful, the case will be closed or reopened depending on whether any additional changes are needed. 
    \item Reopened: Although this issue has already been solved, the solution is either flawed, omitted some important details, or needs some modifications. Issues are classified as assigned or resolved at the Reopened stage. 
    \item Closed: The issue is now regarded as resolved, and the current solution is accurate. Issues that have been closed can be reopened at a later time as needed.
\end{itemize}

 \section{Approach}
 \label{sec:approach}
This section outlines the methodology employed in this study to address the research objectives. It is divided into several sections, each focusing on a crucial step in the process. The section begins with an overview of the data collection, which details the sources and methods used to gather the necessary data for analysis. The preprocessing describes the steps taken to clean and transform the raw data into a format suitable for further analysis. Next, feature extraction explains the techniques used to extract relevant features from the preprocessed data, capturing essential information for the subsequent classification. Finally, the classification discusses the algorithms and models employed to classify the issues based on their assigned labels. 

\subsection{Data Collection}

Our raw data come from the issues of five industrial projects documented on Jira software of a company that offers solutions in the fields of business-to-business (B2B) display systems for televisions, namely, hotel TV systems and advertising and promotional display systems. The company\footnote{https://www.vestel.com} is a home and professional appliance manufacturer with 18 subsidiaries specializing in electronics, large appliances, and information technology and it is Europe's leading television manufacturer, accounting for a quarter of the European market with over eight million units sold in a year.

\begin{table}
\centering
\caption{Data sets used in the experiments.}\label{table:data}
\begin{tabular}{|c|c|c|c|c|}
\hline
\textbf{ID} & \textbf{Name} & \textbf{\# Issues} & \textbf{Timespan} & \textbf{Team Size}\\
\hline
P1 & Ip Hotel Tv & 1287 & 2011-2021 & 35 \\

P2 & Rf Hotel Tv & 2004 & 2017-2021 & 15 \\

P3 & Hospital Tv & 202 & 2017-2021 & 28 \\

P4 & Vsync & 126 & 2017-2021 & 7 \\

P5 & Html Hotel Tv & 1705 & 2018-2021 & 16 \\
\hline
\end{tabular}
\end{table}

Table \ref{table:data} summarizes the raw data. We use issue reports from five web application projects, all of which are two-sided apps with a management panel and a TV-accessible interface. The mission of the IP Hotel TV project is a browser-based interactive hospitality application used by hotel guests in their rooms. There is also a management application for managing the content that will be displayed. This is the company's first hospitality project, which began in 2011 and is still in use today by many hotels. The project Hospital TV is a hospital-specific version of the IP Hotel TV application. It is compatible with the Hospital Information Management System (HIMS), which is an integrated information system for managing all aspects of a hospital's operations, including medical, financial, administrative, legal, and compliance. The Rf Hotel TV project is a version of the Ip Hotel TV application that can be used in non-intranet environments. A coax cable is used for communication with the server. The HTML Hotel TV project is a cutting-edge hospitality platform. It will eventually replace the IP Hotel TV project. Instead of using an intranet, this version is a cloud application that works over the Internet. A central system oversees the management of all customers. Customers now have access to new features such as menu design and theme creation. The project Vsync is a signage application that synchronizes the media content played by televisions. Televisions play the media through their own players.

These projects are developed in different programming languages. The project Rf Hotel TV is written in Python using the Django framework while the project Vsync is written in C\# and JavaScript using the Angular framework. The rest of the projects are written in pure Javascript and C\# using Microsoft technologies such as .Net and .Net Core frameworks. 

The number of issues per project ranges from 126 to 2004, and the total number of issues is 5324. The data set contains issue reports submitted between 2011 and 2021, all related to different versions of the applications. The issues are created by users with different roles such as software developers, team leaders, testers, or even customer support teams in the data. Then, they are assigned to workers with different roles and experiences. The number of employees in the projects varies between seven and 35 when the ``Creator", ``Reporter", and ``Assignee" columns in the data set are combined.

\begin{figure}
  \begin{subfigure}{0.5\linewidth}
    \centering
    \includegraphics[width=\linewidth]{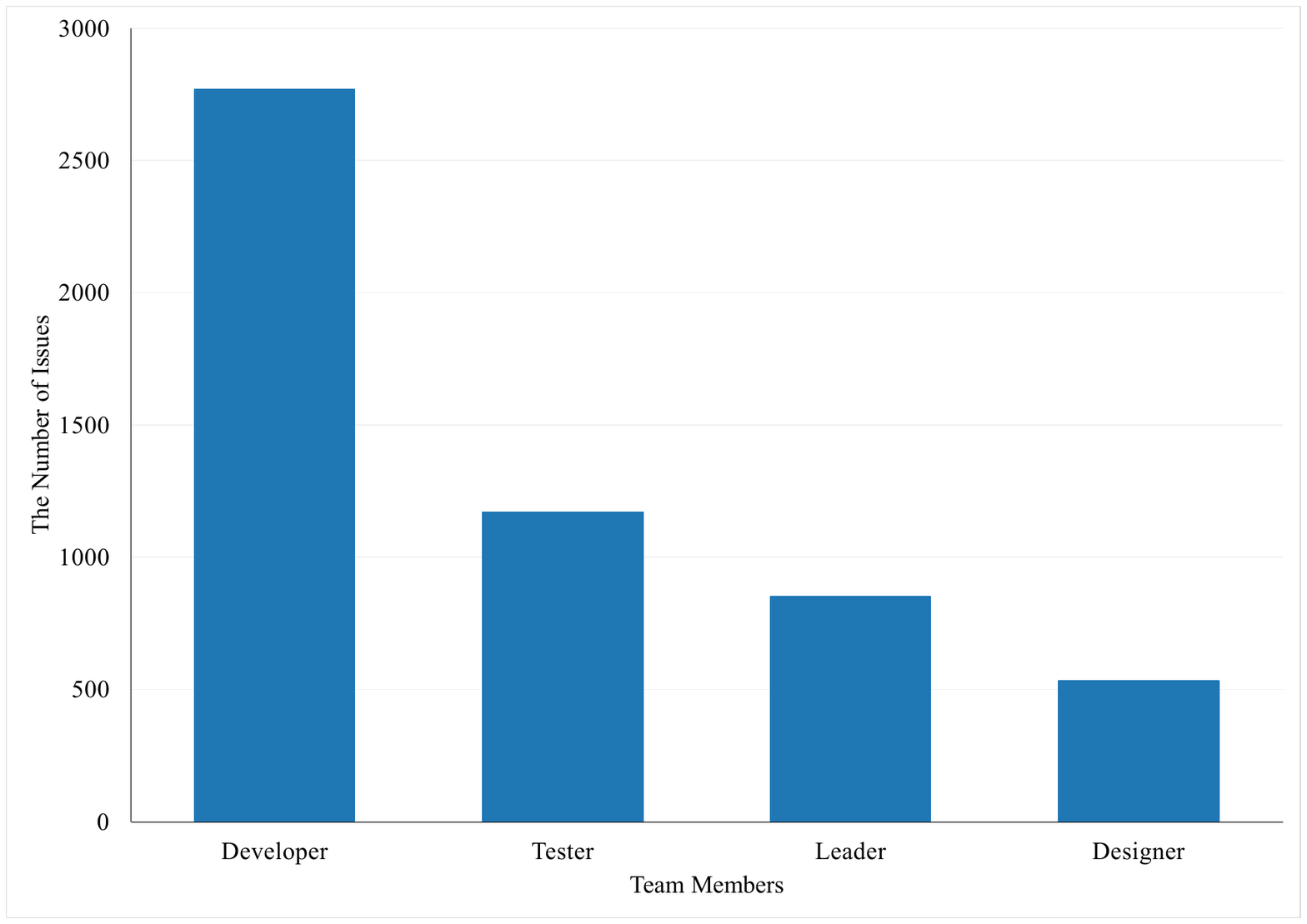}
    \caption{Team members' roles.}
    \label{fig:teammemberdist}
  \end{subfigure}
  \hfill
  \begin{subfigure}{0.5\linewidth}
    \centering
    \includegraphics[width=\linewidth]{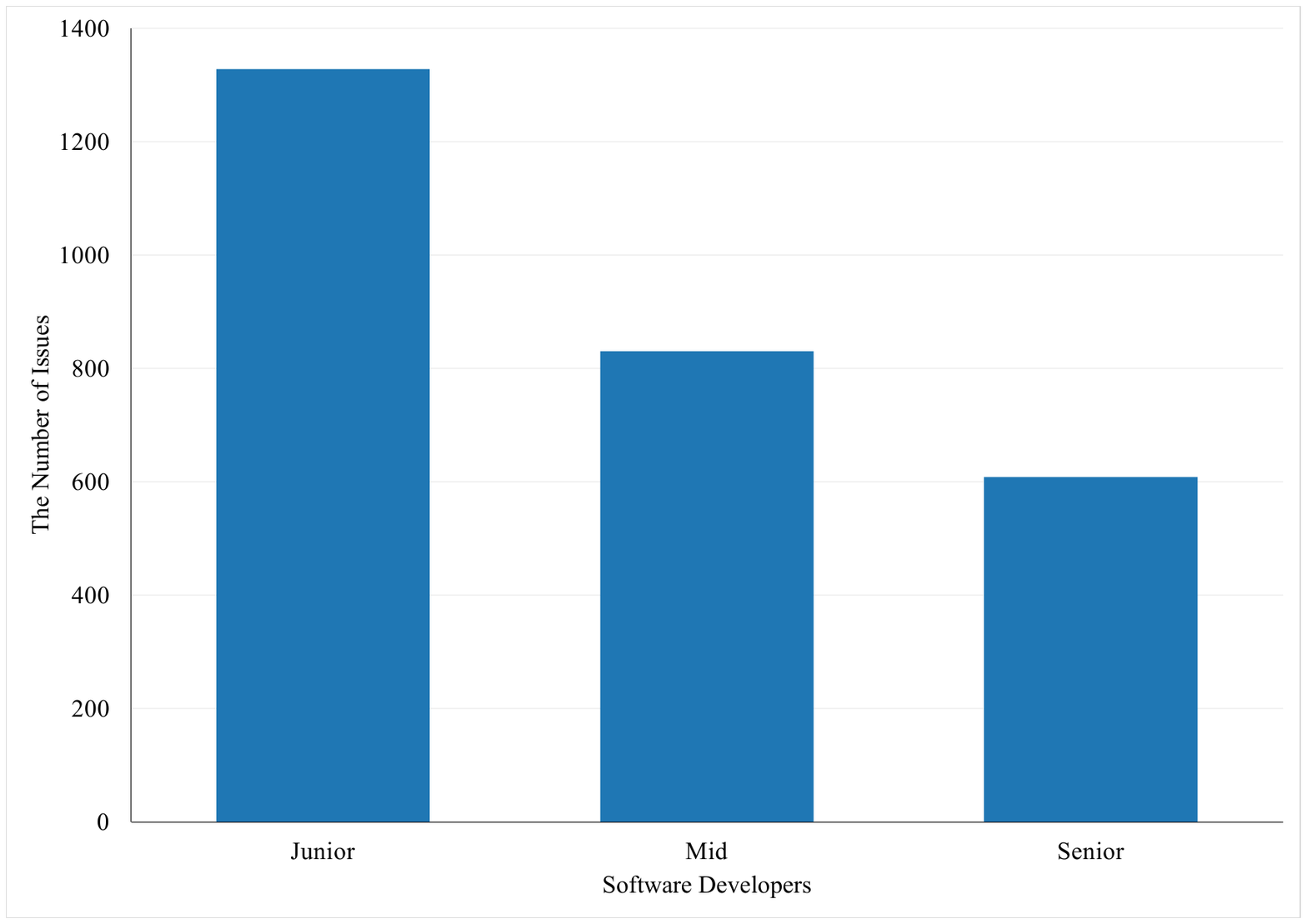}
    \caption{Software developers' experience.}
    \label{fig:developerdist}
  \end{subfigure}
  
  \caption{Distribution of team roles and developers' experience.}
  \label{fig:combined}
\end{figure}

In the original data, the issues are assigned to the individual employees. We removed the names of the individuals to preserve their privacy and inserted their roles in the development team as a new column with the possible values ``Software Developer", ``UI/UX Designer", ``Test Engineer", and ``Team Leader". For the developers only, we also assigned another column indicating their level of expertise as Junior, Mid, and Senior. We inserted this new information in collaboration with the company. Figure \ref{fig:teammemberdist} depicts the distribution of the assignees over the entire data set. As the first chart shows we can observe that team leaders receive the least number of issues and software engineers receive the majority of them. According to Figure \ref{fig:developerdist}, the distribution of experience among software developers, the issues are primarily assigned to junior-level developers, at most, and senior-level developers, at least.

\begin{table}
\centering
\caption{Part of issue from our data set.}\label{table:partOfData}
\begin{tabular}{|c|>{\raggedright\arraybackslash}m{4cm}|m{6.5cm}|}
\hline
\textbf{Project}
& \textbf{Summary}  
& \textbf{Description}\\
\hline
P1 & Orders placed on the room service page do not send e-mails & Although the success message is displayed on the screen, the order e-mail does not come. \\

P2 & Server v1.0.9 Test Request & Please test it. \\

P3 & Making a mother-baby record distinction & The mother's data should come into the room, since it is the same protocol number as the baby, it should be separated. \\

P4 & Multiple video wall synchronization support & Multiple video wall setup should be added to the system and it should be synchronized independently. \\

P5 & Version Filter MacId Problem & Problem experienced due to different incoming Mac address in wifi and ethernet connections. \\
\hline
\end{tabular}
\end{table}

We directly export the data from the company's Jira platform in Excel format, including all columns. Table \ref{table:partOfData} is a small portion of the massive amount of the data available. Although most columns are empty for each row, the tables have a total of 229 columns. To create the issue, required fields like ``Summary," ``Description" and ``Assignee" are filled, but fields like ``Prospect Effort" and ``Customer Approval Target Date" are left blank because they aren't used in the project management. 

The issues are originally written in Turkish with some English terms for technical details and jargon. We also translate the issues to English and share our data publicly in our repository\footnote{https://github.com/busrat/automated-software-issues}. 

\subsection{Preprocessing}
\label{sec:preprocess}

\begin{figure}
\includegraphics[width=\textwidth]{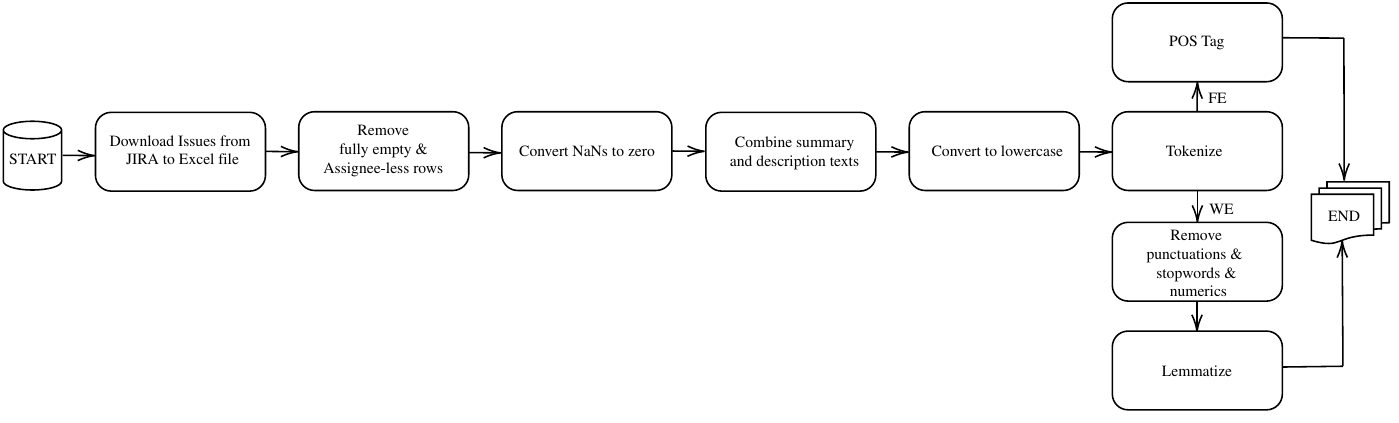}
\caption{Preprocessing steps (FE: Feature Extraction, WE: Word Embedding).}
		\label{fig:preprocessing}
\end{figure}

Preprocessing is the first step in machine learning, and it involves preparing raw data for analysis as shown in Figure \ref{fig:preprocessing}. The process starts with exporting all the issues from the selected projects in the Jira tracker system to an Excel file. Once exported, the data needs to be cleaned up to eliminate any rows that have fully empty columns. We eliminate the rows that contain empty assignee columns. Another issue that needs to be addressed during preprocessing is dealing with missing values. In the Jira tracker system, if the numerical data that will be used as a feature, such as reopen count, is not assigned, it appears as NaN (Not a Number) in the data set. To avoid this problem, the missing values are changed to zero in the entire data set.

We concatenate two textual parts summary and description into new metadata, which we refer to as issue text. Note that these two fields are available for each issue when an issue report is submitted. We apply a lowercase transformation to ensure consistency in the issue text. This step involves converting all uppercase characters to their corresponding lowercase characters. After the transformation, we tokenize the text into words by splitting it into spaces between words.  

For our feature extraction methodology, we do not perform additional text cleaning steps as every word's feature is essential for our process. We perform Part-of-Speech (POS) tagging after the tokenization step. It involves assigning a POS (such as a noun, verb, adjective, etc.) to each word in a given text. We use Zeyrek library \cite{Zeyrek} for issue texts in Turkish because it is trained on a large corpus of Turkish text. 

For most used word embedding methods, we perform additional text cleaning steps to reduce the dimensionality of the data, remove redundant information, and further improve the accuracy. We eliminate all numeric characters and punctuation marks from issue texts. Stop words are words that do not carry significant meaning in a given context and can be safely ignored without sacrificing the overall meaning of a sentence. Examples of stop-words in English include ``the", ``and", ``is", ``are", etc. Similarly, in Turkish, examples of stop-words include ``ve", ``ile", ``ise", ``ama", etc. We use NLTK which provides a built-in list of stop-words for many languages, including Turkish to remove them from issue texts. The last step is lemmatization which is a crucial step in NLP that involves reducing words to their base or dictionary form, known as the ``lemma". The resulting lemma retains the essential meaning of the original word, making it easier to process and analyze text data.

\subsection{Feature Extraction}
This section describes the feature selection steps for the vectors we created and two popular word embedding methods to compare them. The data set obtained from Jira contains over a hundred important fields that we can potentially use as features. However, a significant number of these fields are empty as they are not filled out by the project's team. To avoid this issue, we have narrowed down the selection of features to only those fields that are either non-empty for each issue or automatically populated by the Jira system when an issue is opened.

The columns of the issue tracking system are utilized as the initial feature set in our study, as presented in Table \ref{table:fromjira}. FJN indicates the numerical features from Jira ITS. We consider the data numbers in the columns for these values.
Watchers are the users who have expressed interest in a particular issue and want to receive notifications about its progress. They can receive notifications whenever a comment is added or any changes are made to the issue. Typically, multiple individuals subscribe to problematic issues in order to receive notifications upon closure or new comments. Images column is used to attach relevant screenshots, diagrams, or other images to an issue. This helps in better understanding and resolving the issue. When a bug cannot be easily identified or located, it is common practice for test engineers to include an image of the bug as a reference for developers. This serves as a visual aid to help the developers understand the issue and resolve it more effectively. Reopen Count column tracks the number of times an issue has been reopened after being marked as resolved or closed. It provides insight into the recurring nature of the issue and can help identify if the issue is resolved properly or not. This feature serves to distinguish problematic issues that persist even after the developer has addressed them. Reassign Count column keeps track of how many times an issue has been reassigned to different users or teams. It can help in analyzing the workflow and identifying any inefficiencies. There are various reasons why an issue may be assigned to someone other than the initially assigned individual. These reasons include cases where the assigned person is unavailable or unable to resolve the issue. 
The linked issues column allows users to link related issues together. It helps in identifying dependencies and tracking progress across multiple issues. The sub-tasks column allows users to break down larger issues into smaller sub-tasks. It helps in better managing and tracking complex issues. The components column specifies the different modules or components of the software that are affected by the issue. It helps in identifying the root cause of the issue and assigning it to the appropriate team or individual. 

We only consider whether or not there is a value present in the column for columns that are mostly empty across the issues and do not have diversity in the data to separate each other. We call these boolean features FJB. Reported by customer column indicates if a customer or an internal team member reports the issue. It helps in prioritizing and resolving customer-reported issues quickly. The tested versions column indicates the versions of the software in which the issue is tested. It helps in identifying the specific version where the issue is first detected. The test execution type column specifies the type of test execution, such as Manual or Automated. It helps in tracking the progress and success of the testing phase. The approval type column is used to indicate the type of approval required for the issue, such as Manager Approval or Technical Approval. It helps ensure that the issue is reviewed and approved by the appropriate stakeholders before being resolved. Affects versions column indicates the versions of the software that are affected by the issue. It helps in identifying the scope of the issue and prioritizing it accordingly. 

Several features in our feature set are categorical as FJC, and in order to use them in our analysis, we replaced them with numerical values using label encoding. This process assigns a unique numerical value between 0 and the number of classes minus one to each category, allowing us to use them in our computations. The issue type column defines the type of issue being reported, such as Bug, Improvement, Task, etc. It helps in categorizing and prioritizing issues based on their type. The reporter column indicates the user who reported the issue. It can help in contacting the user for additional information or to gather feedback. The priority column indicates the relative importance of an issue. It can be set to High, Medium, Low, or any other custom value based on the severity of the issue and its impact on the project. The frequency column tracks how often the issue occurs. It helps in identifying patterns and trends in the occurrence of the issue. The bug category column allows users to categorize the issue based on its root cause, such as Performance, Security, Usability, etc. It helps in prioritizing and assigning the issue to the appropriate team or individual. The labels column allows users to add descriptive tags to an issue. It helps in categorizing and searching for issues based on common themes or topics. 

Issue texts are utilized to extract features using NLP techniques, as detailed in Table \ref{table:fromtext}. The FTN column indicates the numerical features extracted from the text fields. The Summary Words and Description Words columns indicate the number of words in the corresponding issue text columns. To analyze the sentiments of the issue texts, the TextBlob library \cite{loria2018textblob} is used for sentiment analysis. Polarity represents the emotional tone of a piece of text, typically characterized as positive, negative, or neutral. The polarity score ranges from minus one (most negative) to one (most positive). Subjectivity, on the other hand, refers to the degree to which a piece of text expresses opinions or personal feelings, as opposed to being factual or objective. The subjectivity score ranges from zero (most objective) to one (most subjective). As described in Section \ref{sec:preprocess}, each word in the issue text is classified with known lexical categories using POS tagging for the morpheme-related features in both Turkish and English. The number of available tags, such as adjective, adverb, conjunction, verb, numeral, etc., is added as a new feature column for each issue. However, not all tags are added to the table. The most effective POS tags as features are discussed in Section \ref{section:feature_comparison}. The FTB column indicates the boolean features extracted from the text fields. The issue text is searched for Bug Words, such as ``error", ``null", ``bug", ``server", and ``undefined" to determine if there is a bug or for the developers. Test Words, such as ``test" and ``request" are searched for issues created for the test engineers. Document Words, such as ``document(ation)" and ``write" are searched for the team leaders, and Design Words, such as ``design", ``icon", and ``logo" are searched for the designers. The negative verb is a boolean value that checks for negative verbs in the issue text. It is assumed that bugs would be more likely to have negative verbs in their definitions rather than being by design or a new task opened. The necessity verb is a boolean value that checks for the verb for necessity in the issue text (e.g., ``should" verb in English, ``-meli/-malı" suffix in Turkish).

\begin{table}
\centering
 \caption{Features from the columns of the issue tracking system.}
\begin{tabular}{|c|c|p{7.5cm}|}
\hline
\textbf{Feature} & \textbf{Name} & \textbf{Description} 
  \\\hline
FJN1 & Watchers & The number of users following the issue. 
\\
FJN2 & Images & The number of the images that have been attached to the issue. 
\\
FJN3 & ReopenCount & The number of times an issue has been reopened. 
\\
FJN4 & ReassignCount & The number of times an issue has been reassigned to different users.
\\
FJN5 & LinkedIssues & The number of linked related issue keys to the issue. (i.e. ProjectName-2037) 
\\
FJN6 & SubTasks & The number of added sub-issue keys to the issue. (i.e. ProjectName-2037)
\\
FJN7 & Components & The number of different components of the software that are affected by the issue (i.e. cloud)
\\
FJB1 & ReportedByCustomer & The customer who reports the issue. (i.e X Hotel) 
\\
FJB2 & TestedVersions & The tested versions of the software in which the issue is tested. (i.e. ProjectName 9.4.x) 
\\
FJB3 & TestExecutionType & The type of test execution (i.e. manual) 
\\
FJB4 & ApprovalType & The type of approval required for the issue. (i.e. P1-Pilot) 
\\
FJB5 & AffectsVersions & The versions of the software that are affected by the issue. (i.e. ProjectName 9.4.x) 
\\
FJC1 & IssueType & The type of issue. (Story, Epic, Request, Bug, Test Request, Technical task)
\\
FJC2 & Reporter & The user who reported the issue. 
\\
FJC3 & Priority & The relative importance of an issue. (Low, Medium, High, Showstopper) 
\\
FJC4 & Frequency & The frequency of the issue occurs. (i.e. always-2/2, sometimes-2/4) 
\\
FJC5 & BugCategory & Category of the issue based on its root cause (i.e. General, Functional) 
\\
FJC6 & Labels & Added descriptive tags to an issue. (i.e. Admin) 
\\\hline
\end{tabular}
\label{table:fromjira}
\end{table}

Word Embedding techniques are used to represent text as vectors. To create vectors, we utilize the preprocessed combination of title and description parts of issues. There are various forms of word embeddings available, with some of the most popular ones being Bag of Words (BoW) \cite{bow1996}, Term Frequency-Inverse Document Frequency (TF-IDF) \cite{tfIdf2004} and, Word2Vec \cite{Word2Vec}. We have implemented Tf-Idf and BOW algorithms using the Sklearn library \cite{scikit-learn} and the Word2Vec algorithm using the Gensim library \cite{rehurek2011gensim}. We have tested both BoW unigram and bigram models separately and together. The unigram model stores the text as individual tokens, while the bigram model stores the text as pairs of adjacent tokens. Based on our experiments, the BoW unigram model outperformed the bigram model. This is attributed to the unigram model's superior ability to capture essential text features. 

\subsection{Classification}
We can train a classifier to attempt to predict the labels of the issues after we have our features. We experiment with various algorithms and techniques when working on a supervised machine learning problem with a given data set in order to find models that produce general hypotheses, which then make the most precise predictions about future instances, possible. We start with using machine learning techniques that Scikit-learn includes several variants of them to automatically assign issue reports to the developers. We try the best-known ML models i.e. Support Vector Machine (SVM), Decision Tree, Random Forest, Logistic Regression, k-nearest Neighbors (kNN), and Naive Bayes (NB). We use Multinomial and Gaussian NB which are the most suitable variants for text classification. The multinomial model offers the capability of classifying data that cannot be numerically represented. The complexity is significantly decreased, which is its main benefit. We test the one-vs-rest model with SVM, a heuristic technique for multi-class binary classification algorithms. The multi-class data set is divided into various binary classification issues. 
Scikit-learn offers a high-level component called CountVectorizer that will produce feature vectors for us. The work of tokenizing and counting is done by CountVectorizer, while the data is normalized by TfidfTransformer. In order to combine this tool with other machine learning models, we supply the title and description fields that we combined. 

Most machine learning algorithms do not produce optimal results if their parameters are not properly tuned so we use grid search with cross-validation to build a high-accuracy classification model. We use the GridSearchCV tool from Sklearn library \cite{scikit-learn} to perform hyperparameter tuning in order to determine the optimal values for a given model. In particular, we use a 10-fold cross-validation. We first split the issues data set into 10 subsets.  We train the classifier on nine of them and one subset is used as testing data. Several hyper-parameter combinations are entered, then we calculate the accuracy and the one with the best cross-validation accuracy is chosen and used to train a classification method on the entire data set.

We also try ensemble learning methods \cite{Ensemble} which combine the results of multiple machine learning algorithms to produce weak predictive results based on features extracted from a variety of data projections, and then fuse them with various voting mechanisms to achieve better results than any individual algorithm. First, we use the hard-voting classifier which can combine the predictions of each classifier to determine which class has the most votes. Soft voting based on the probabilities of all the predictions made by different classifiers is also an option. Second, we try a classification method called extra trees, which combines the predictions of multiple decision trees. Finally, we combine machine learning methods with bagging, boosting, and stacking ensemble learning techniques. While boosting and stacking aim to create ensemble models that are less biased than their components, bagging will primarily focus on obtaining an ensemble model with less variance than its components \cite{odegua2019empirical}. 

\begin{table}
\centering
 \caption{Features extracted from issue texts.}
\begin{tabular}{|c|c|p{8cm}|}
\hline
\textbf{Feature} & \textbf{Name} & \textbf{Description} 
\\\hline
FTN1 & SummaryWords & The number of words in the issue summary. 
\\
FTN2 & DescriptionWords & The number of words in the issue description. 
\\
FTN3 & PolarityScore & Emotional tone of the issue text. (ranges from minus one (most negative) to one (most positive)) 
\\
FTN4 & SubjectivityScore & Issue text expresses opinions or personal feelings. (ranges from zero (most objective) to one (most subjective)) 
\\
FTNP & PosTags & The number of every POS tag in the issue text. 
\\
FTB1 & BugWords & Check for ``error, null, bug, server, undefined" words in the issue text. 
\\
FTB2 & TestWords & Check for ``test, request" words in the issue text. 
\\
FTB3 & DocumentWords & Check for ``document/ation, write" words in the issue text. 
\\
FTB4 & DesignWords & Check for ``design, icon, logo" words in the issue text. 
\\
FTB5 & NecessityVerb & The boolean value that checks for a verb for necessity in the issue text. (i.e. ``should" verb in English, ``-meli/-malı" suffix in Turkish) 
\\
FTB6 & NegativeVerb & The boolean value that checks for negative verbs in the issue text. (i.e. ``n't, not" in English, ``-me, -ma" in Turkish) 
\\\hline
\end{tabular}
\label{table:fromtext}
\end{table}

The majority of classification studies using the issue data set do not use or have limited success with deep learning-based text mining techniques. Herbold et al. \cite{Herbold2020} believe that they lack sufficient (validated) data to train a deep neural network and deep learning should instead be used for this task once the necessary data requirements are satisfied, such as through pre-trained word embeddings based on all issues reported at GitHub. We try some bidirectional language models: DistilBert, Roberta, and Electra to provide empirical evidence. DistilBert \cite{DistilBert} is developed using the Bert \cite{Bert} model. In comparison to pre-trained Bert on the same corpus, this model is quicker and smaller in size. Roberta \cite{Roberta} is retraining BERT with improved training methodology, more data and compute power. Electra \cite{Electra} uses less computation than Bert to pre-train transformer networks. In 2022, Guven \cite{Guven2022} compares language models for the Turkish sentiment analysis approach and the best performance has been achieved by training the Electra language model. These models are pre-trained with a Turkish data set for Turkish approaches \cite{Stefan2020}.

\section{Experiments and Results}
\label{sec:results}
This section presents the findings and outcomes of the conducted experiments, providing a comprehensive analysis of the collected data. We critically assess the performance and effectiveness of the proposed methodology by employing various evaluation metrics and techniques. We compare the extracted features from different perspectives, evaluating their individual contributions to the classification task. Lastly, we present a thorough statistical examination of the obtained results, employing appropriate statistical tests and measures to validate the significance and reliability of the findings. 

\subsection{Evaluation}
Table \ref{table:results_assignment} presents the experiment results for the issue assignment. The models are evaluated on Team Assignment (TA) and Developer Assignment (DA). The stacking model (RF and Linear SVC) achieved the highest accuracy, with values of 0.92 for TA and 0.89 for DA. Other models, such as Support Vector Machine, Logistic Regression, and Random Forest, also showed good performance with accuracies ranging from 0.86 to 0.88. The transformer-based models, including DistilBert, Roberta, and Electra, demonstrated competitive accuracies, with Roberta and Electra achieving the highest scores in some cases.

\begin{table}
\centering
 \caption{Experiment results (TA: Team Assignment, DA: Developer Assignment).}
\begin{tabular}{|c|c|c|}
\hline
 \textbf{Classification Model} & 
 \textbf{Acc\_TA} & 
 \textbf{Acc\_DA} 
\\\hline
 Support Vector Machine & 0.88 & 0.86 
 \\
Logistic Regression & 0.88 & 0.85 
\\
Naive Bayes & 0.83 & 0.78 
\\
Multilayer & 0.86 & 0.75 
\\
Stochastic Gradient Descent & 0.87 & 0.82 
\\
Decision Tree & 0.86 & 0.86 
\\
Random Forest & 0.88 & 0.86 
\\
KNN & 0.88 & 0.88 
\\
One vs Rest & 0.82 & 0.81 
\\
Voting Soft & 0.87 & 0.87 
\\
Voting Hard & 0.88 & 0.88 
\\
RF with Boosting & 0.90 & 0.88 
\\
Bagged DT & 0.86 & 0.88 
\\
Extra Trees & 0.89 & 0.88 
\\
Stacking (RF and Linear SVC) & \textbf{0.92} & \textbf{0.89} 
\\
DistilBert & 0.88 & 0.87 
\\
Roberta & 0.91 & 0.88 
\\
Electra & 0.91 & 0.88 
\\\hline
\end{tabular}
\label{table:results_assignment}
\end{table}

\begin{table}
\centering
 \caption{Performance metrics for each class in the Stacking algorithm.}
\begin{tabular}{|c|c|c|c|c|}
\hline
\textbf{Task} & 
 \textbf{Class} & 
 \textbf{Precision} & 
 \textbf{Recall} & 
 \textbf{F1}
\\\hline
 \multirow{4}{*}{TA} & Developer & 0.90 & 0.99 & 0.94 
 \\
& Tester & \textbf{0.95} & \textbf{0.71} & \textbf{0.83}
 \\
& Designer & 0.92 & 0.67 & 0.80 
 \\
& Leader & 0.65 & 0.57 & 0.62 
\\\hline
\multirow{3}{*}{DA} & Senior & 0.90 & 0.67 & 0.80 
 \\
& Mid & 0.89 & 0.94 & 0.91 
 \\
& Junior & 0.62 & 0.53 & 0.57 
\\\hline
\end{tabular}
\label{table:results_classes_IA}
\end{table}

Table \ref{table:results_classes_IA} provides the performance metrics for each class in the Stacking algorithm for issue assignment. Under the Team Assignment (TA) approach, the Stacking algorithm achieved a high precision value of 0.90 for the Developer class, indicating a low rate of false positive assignments. The Recall score of 0.99 for the Developer class demonstrates the algorithm's ability to correctly identify the majority of instances assigned to developers. The Tester class shows a balance between precision (0.95) and recall (0.71), indicating accurate assignments with a relatively high rate of false negatives. The Designer class exhibits similar trends with a precision of 0.92 and a recall of 0.67. The Leader class has relatively lower precision and recall scores, indicating more challenging assignments for the algorithm.

Under the Developer Assignment (DA) approach, the Stacking algorithm achieved high precision values for the Senior class (0.90) and the Mid class (0.89), indicating accurate assignments with low rates of false positives. The Mid class also demonstrates a high recall score of 0.94, indicating effective identification of instances assigned to this class. The Junior class shows a lower precision (0.62) and recall (0.53) compared to the other classes, suggesting potential challenges in accurately assigning instances to this class.

\begin{figure}
\includegraphics[width=\textwidth]{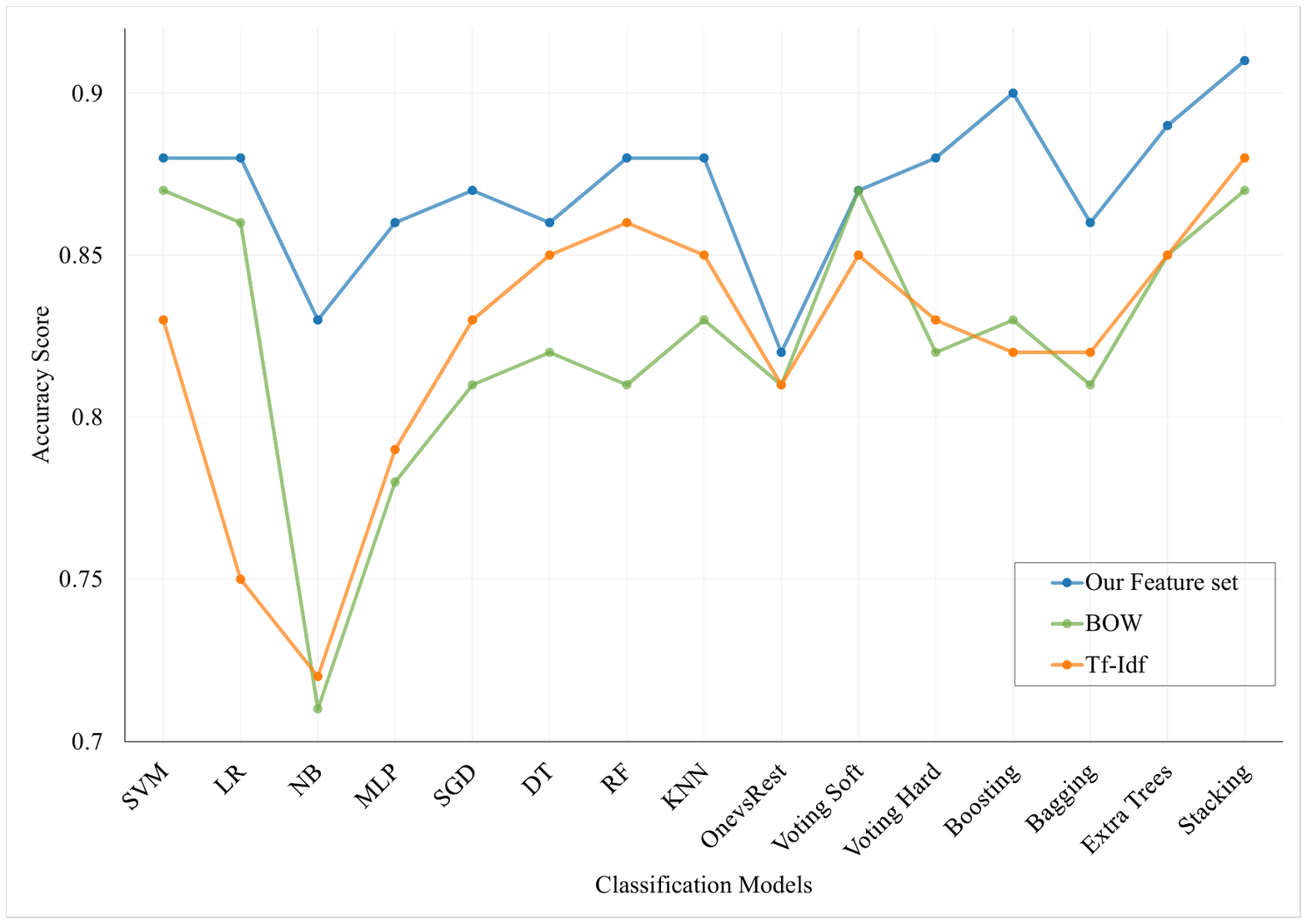}
\caption{Comparison of our feature set with word embedding methods.}
		\label{fig:line_assignment}
\end{figure}

We also test our classification algorithms using the most popular word embedding techniques to determine how well our features work. Figure \ref{fig:line_assignment} illustrates the comparison of our feature set with Tf-Idf and BOW methods for the issue assignment. Despite the potential of Word2Vec as a word embedding algorithm, the accuracy results in my approach do not yield comparable outcomes. We use the accuracy score as the comparison metric. The graph demonstrates that using our feature set yields superior results while using Tf-Idf and BOW yields comparable results.

\subsection{Feature Comparison}
\label{section:feature_comparison}
\begin{figure}
\includegraphics[width=\textwidth]{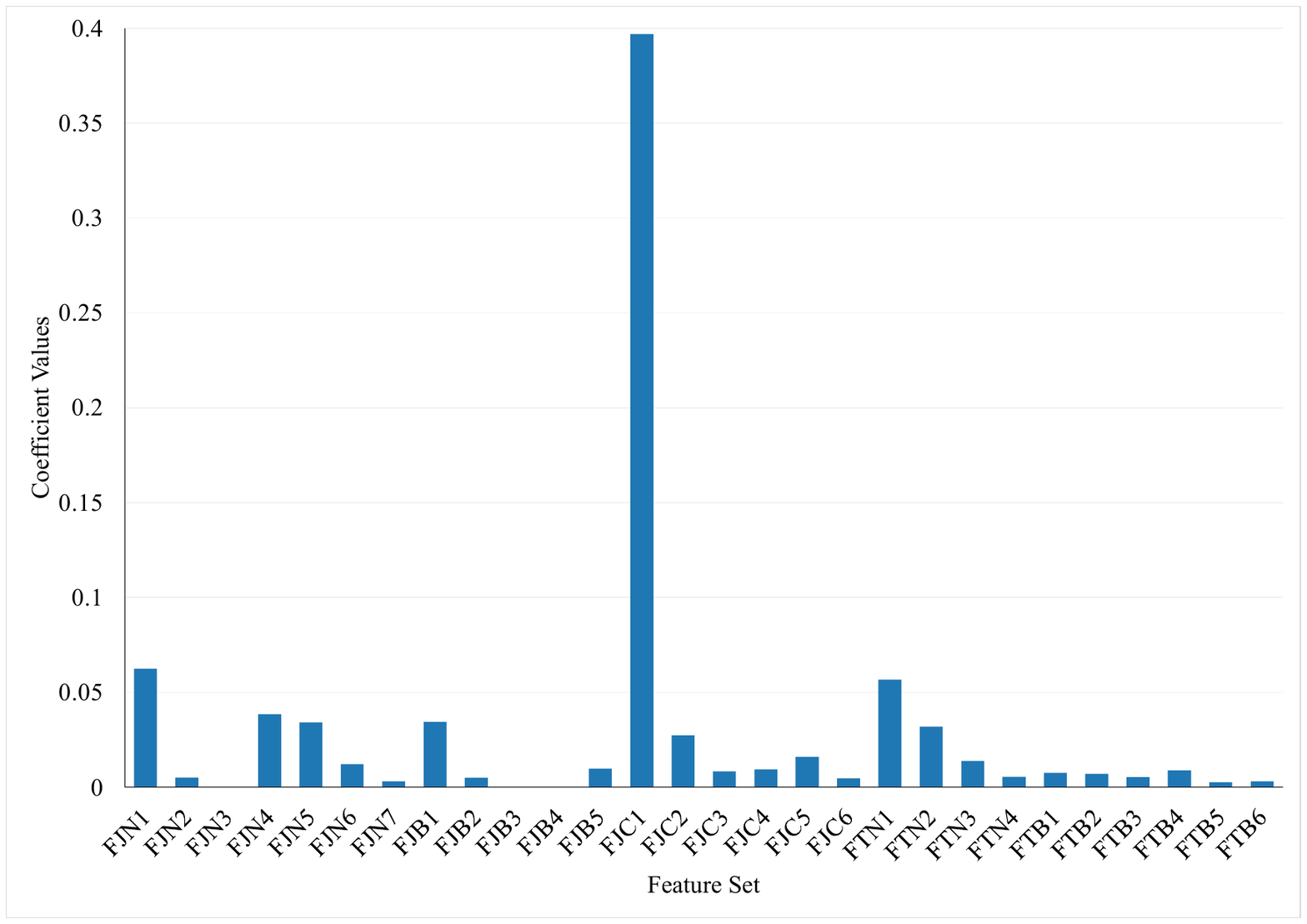}
\caption{Coefficient values of each feature.}		
  \label{fig:issue_assignment_feature}
\end{figure}

Reducing the number of redundant and irrelevant features is an effective way to improve the running time and generalization capability of a learning algorithm \cite{Dash1997feature}. Feature selection methods are used to choose a subset of relevant features that contribute to the intended concept. These methods can employ a variety of models and techniques to calculate feature importance scores. One simple approach \cite{Brownlee2019choose} is to calculate the coefficient statistics between each feature and the target variable. This method can help to identify the most important features of a given problem and discard the redundant or irrelevant ones. By reducing the number of features used for training a model, the running time of the algorithm can be significantly reduced without sacrificing accuracy. Moreover, feature selection can also improve the interpretability of the model, as it helps to identify the key factors that influence the target variable. We present the coefficient values of each feature in Figure \ref{fig:issue_assignment_feature}.

\begin{figure}
\includegraphics[width=\textwidth]{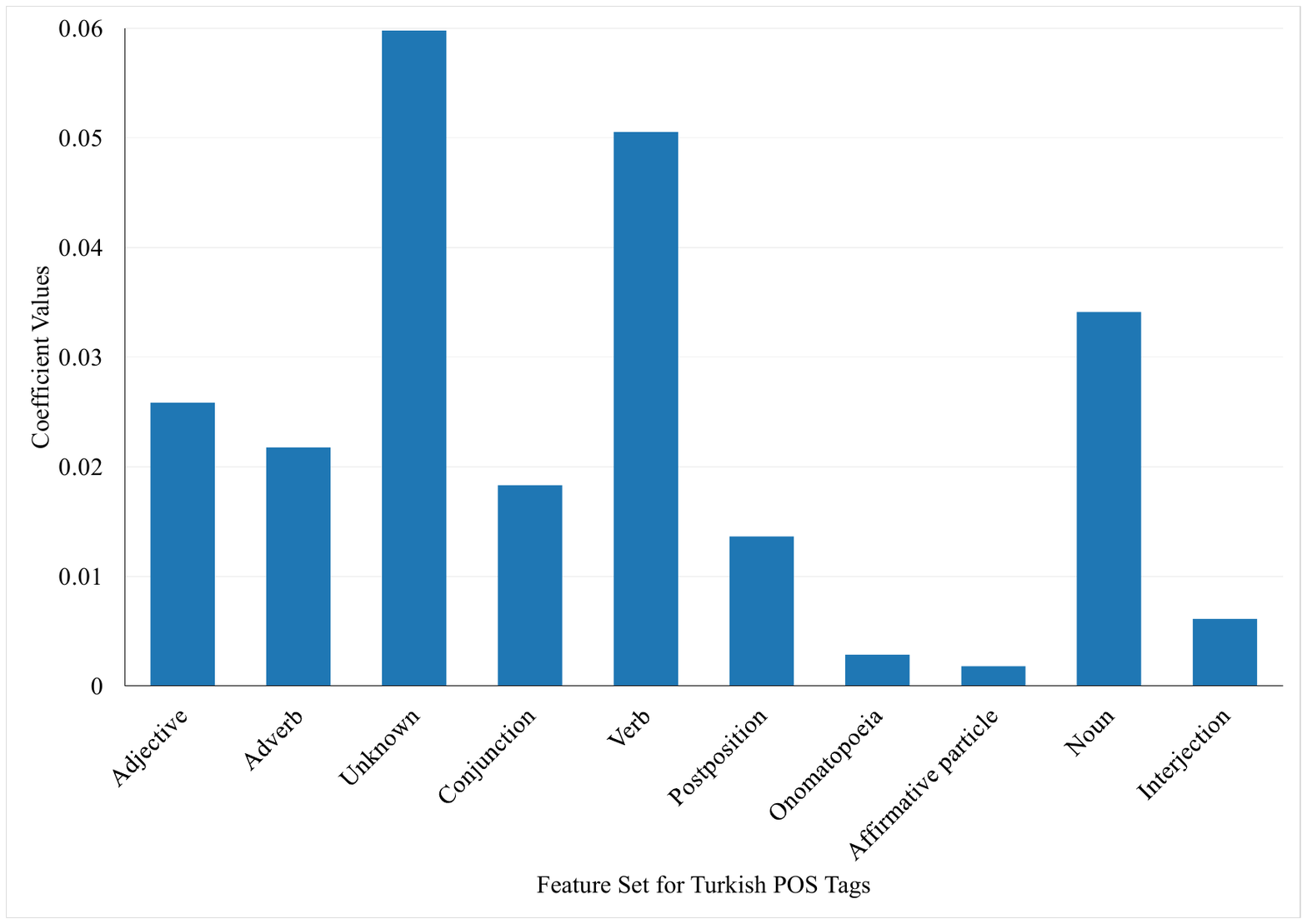}
\caption{Coefficient values of each POS tag in Turkish.}
		\label{fig:postags_turkish}	
\end{figure}

We find that Issue Type, namely FJC1, emerges as the most influential feature from our feature set. Apart from the Issue Type, we discover that the features Watchers and Summary Words also exhibit significant effectiveness in our analysis. Conversely, features such as Reopen Count, Test Execution Type, and Approval Type demonstrate no impact on our issue assignment process. In Figure \ref{fig:postags_turkish}, we present the effective POS tags in Turkish, highlighting the most influential ones among all POS tags. Notably, the number of unknown words, verbs, and nouns emerge as the most impactful features. Following the rigorous selection of the best features, we proceed to employ Scikit-learn's \cite{scikit-learn} SelectFromModel, a powerful meta-transformer designed to choose features based on their importance weights, to retrain our models. Through this process, we carefully identify and select the top eight features that exhibited the highest significance, as determined by the module. Remarkably, leveraging this refined feature subset allows us to achieve optimal performance and attain the most favorable outcome in our experiments.

\subsection{Statistical Analysis}
\label{sec:statistical}
 
Statistical significance tests are conducted to compare classification methods for determining whether one learning algorithm outperforms another on a particular learning task. Dietterich \cite{Dietterich1998} reviews five approximate statistical tests and concludes that McNemar's test and the 5x2 cv t-test, both have low type I error and sufficient power. In our study, we combine all data sets into a single data set for the classification algorithm. Dietterich \cite{Dietterich1998} recommends using a 5x2 t-test to statistically compare two classifiers on a single data set. The 5x2 f-test, which is also suggested as the new standard by the original authors above, is further expanded upon by Alpaydın \cite{Alpaydin1999}. The following lists the null hypothesis and the alternative hypothesis. The null hypothesis is that the probabilities are the same, or in simpler terms, neither of the two models outperforms the other. The alternative hypothesis, therefore, holds that the performances of the two models are not equivalent.

\begin{table}
\centering
 \caption{Analysis of the Stacking algorithm's statistical test results in comparison to others.}
\begin{tabular}{|c|c|c|}
\hline
 \textbf{Classification Model} & 
 \textbf{p-value} & 
 \textbf{Hypothesis}
\\\hline
SVM & 0.0076 & \textbf{Reject} 
\\
Logistic Regression & 0.0412 & \textbf{Reject} 
\\
Naive Bayes & 0.0413 & \textbf{Reject} 
\\
Multilayer & 0.1336 & Accept 
\\
SGD & 0.0412 & \textbf{Reject} 
\\
Decision Tree & 0.0265 & \textbf{Reject} 
\\
Random Forest & 0.4795 & Accept 
\\
KNN  & 0.0412 & \textbf{Reject} 
\\
One vs Rest  & 0.0025 & \textbf{Reject} 
\\
Voting Soft & 0.0736 & Accept  
\\
Voting Hard & 0.1336 & Accept  
\\
RF with Boosting & 0.4795 & Accept  
\\
Bagged DT & 0.4795 & Accept  
\\
Extra Trees & 0.2482 & Accept  
\\
DistilBert & 0.1376 & Accept 
\\
Roberta & 0.3675 & Accept  
\\
Electra & 0.3675 & Accept  
\\\hline
\end{tabular}
\label{table:results_statistical}
\end{table}

Accordingly, we apply the 5x2 f-test implemented by Alpaydın \cite{Alpaydin1999} which is an extension of the 5x2 cv t-test as stated above. We create the matrix for all pairwise comparisons of learning algorithms. In this test, the splitting process (50\% training and 50\% test data) is repeated five times. A and B are fitted to the training split and their performance on the test split in each of the five iterations is assessed. The training and test sets are then rotated (the training set becomes the test set, and vice versa), and the performance is computed again, yielding two performance difference measures. Then, the mean and variance of the differences are estimated and the f-statistic proposed by Alpaydın is calculated as
\begin{equation}
f = \frac{
\sum_{i=1}^{5}\sum_{j=1}^{2}(p_i^j)^2}{2\sum_{i=1}^{5}s_i^2},
\end{equation}
where ${p_i^j}$ is the difference in error rates between the two classifiers on fold $j = \{1,2\}$ of replication $i = \{1, ..., 5\}$ and $s_i^2$ is estimated variance. We reject the null hypothesis that the two models' performances are equal if the p-value is less than our chosen significance level ($p\text{-value} < \alpha = 0.05$) and accept that the two models are significantly different. Table \ref{table:results_statistical} presents the results of the statistical F test analysis comparing the Stacking algorithm to other classification models. The table provides the p-values and corresponding hypothesis decisions for each classification model. Based on the p-values compared to the chosen alpha value, we can accept the null hypothesis that there are no significant differences among all ensemble learning and deep-learning techniques. However, when comparing these techniques with machine learning methods, the majority of cases result in rejecting the null hypothesis, indicating significant performance variations. This observation is evident from the table, which highlights the substantial rejection of the null hypothesis in most comparisons between ensemble learning, deep learning, and machine learning methods.

\section{Discussion and Qualitative Analysis}
\label{sec:discuss}
In this section, we focus on the validity of threats that could impact the reliability and generalizability of our study results. We discuss potential sources of bias, confounding variables, and other factors that may affect the validity of our study design, data collection, and analysis. We also describe our experiment for user evaluation in the company, which is aimed at investigating the effectiveness of our approach for issue assignment. We explain the methodology we use to gather feedback from users, such as surveys or interviews, and how we plan to analyze the results. 

\subsection{Threats to Validity}
In this section, we discuss the validity threats to our study concerning internal validity, external validity, construct validity, and conclusion validity. (Wohlin et al. \cite{WohlinValidity}) 

Internal validity pertains to the validity of results internal to a study. It focuses on the structure of a study and the accuracy of the conclusions drawn. To avoid creating a data set with inaccurate or misleading information for the classification, the corporate employees labeled the employees by fields in the data set. We attempt to use well-known machine learning libraries during the implementation phase to prevent introducing an internal threat that can be brought on by implementation mistakes. All of our classification techniques specifically make use of the Python Sklearn \cite{Shokripour2015} package. Sklearn and NLTK are used to preprocess the text of the issues and Sklearn metrics are used to determine accuracy, precision, and recall. We think that the application that can cause the most internal threat is when allocating Turkish issues to part of speech tags. Since Turkish is not as common as English and is an agglutinative language, it is more difficult to find a highly trained POS tagger library that provides high precision. We decided to use the Turkish\_pos\_tagger \cite{TurkishPosTagger} library by comparing many parameters such as data numbers, accuracy percentages, and usage popularity among many Turkish POS tagger libraries. Turkish\_pos\_tagger library includes 5110 sentences and the data set originally belongs to Turkish UD treebank. For 10-fold validation, the accuracy of the model is 95\%.  

External validity involves the extent to which the results of a study can be applied beyond the sample. We use the data set of five different applications with thousands of issues. These projects use different software languages and various technologies at the front-end and back-end layers, including restful services, communication protocols such as gRPC and WebSocket, database systems, and hosting servers. The issue reports cover a long period of time from 2011 to 2021. However, all the projects we get from the issue reports are mainly concerned with the development of web projects made to run in the browser on the TV. Issues contain many TV-specific expressions, such as application behaviors that occur as a result of pressing a button on the remote or resetting the TV. We make great efforts to ensure that the features we design to prevent external validity concerns are not particular to the data we utilize. For our classification analysis to be applicable in other fields, we believe that it will be sufficient to replicate it using other data sets from various fields.

Construct validity refers to the degree to which a test or experiment effectively supports its claims. The performance of our automated issue assignment system is evaluated using the well-known accuracy metric. We additionally back it up with two other well-known metrics, namely recall, and precision. Tasks in the organization where we use the data set can only be given to one employee, and that employee is also in charge of the sub-tasks that make up the task. This makes assigning the issue report to a single employee group as a binary classification, an appropriate classification method. However, it could be necessary for a business analyst, a product owner, and a software developer to open the same task in different project management or different software teams. For this kind of data set, the binary classification research we conducted is not a suitable approach.    

Conclusion validity refers to the extent to which our conclusion is considered credible and accurately reflects the results of our study. All the issue data we used are real issue reports collected from the software development team. We use issue reports in 10 years time span, but according to the information we received from within the company, the turnover rate is low compared to other companies, and especially the team leaders and testers, who usually create the tasks, are generally people who worked for the company for 10+ years. This may have caused a high similarity in the language, namely the text, of the opened tasks and created a conclusion threat at the accuracy rate. To assess how well the accuracy values we find are consistent among themselves, we used statistical significance tests as outlined in Section \ref{sec:statistical}. By proving our hypotheses in this manner, we showed the consistency of the outcomes we discovered and the effectiveness of our methods. 

\subsection{User Evaluation}
\label{sec:userevaluation}
An employee of the company who served as a Senior Software Developer and created the architecture of the projects where we use the data set as well as an employee who serves as the Team Leader of the three projects we have are interviewed for this section about our application. On all the projects where we use the data set, a software developer has been employed by this company for three years. Three of the projects are being led by the team leader, who has been employed by the company for ten years. After spending about 15 minutes outlining our application, we evaluate the outcomes by conducting a few joint experiments.
In the first section, we find an issue that has been done and our model assigns a different assignee value than the one made on ITS, and we talk about it. This issue is assigned to the team leader in ITS, but our model assigns it to the junior software developer. We want to know what team members think about this example of a wrong assignment. Our model assigns it to a different employee, both in terms of experience and field. First off, they state that from the test team or the customer support team, an incorrectly tested issue that is not actually a problem or issues that should not be developed can be opened. In this case, the team manager can take the issue and bring it to Won't Fix status. This is also the case in this issue. In fact, this is something that should not be done. They state that for such situations, the team manager must decide.

In the second part, we assign an idle issue that has not yet been assigned by our classification method. The model labels the issue as the junior software developer. We are asking for their opinion to find out if this is the correct assignment. Considering the scope of the job, both team members state that it is appropriate to assign this job to a junior friend, as the requirement for seniority is quite low. Assigning it to mid or senior employees would not be a problem either, but they would not consider assigning this issue to more experienced employees.

In the next section, we give a data set consisting of 20 issues assigned and closed in ITS to the senior software developer and team leader in the company. They label these issues according to the labels we set. We compare the tags made with the assigned values in the issue's data set and the assignments made by our best working system. Table \ref{table:comparison} shows the results of this comparison.

First, we compare the labels of Senior Software Developer and Team Leader with the assigned values on the ITS and the label values of our best model and find the issues where all four are the same. The least number of intersections are 11 common labels with values where all four of them are the same. In order to understand whether this difference is due to mislabeling of our model or due to labeling differences between employees and ITS, in combination 2, we check the values where the labels made by our model and all three of the labels made by the employees intersect. Here the total intersection turns out to be 13. We show the labels that these two issues are assigned differently on ITS to the employees. Both employees gave the same tag value, but a different employee type seems to have closed the issue in ITS. They think that if it's a problem that an employee has dealt with before, they may have taken it for that reason and it could be both types of labels. In the third combination, we find the values that our model has labeled in common with at least one employee to see if there are labels that they think differently among the employees or if our model has assigned completely different assignments from the two. Here, the number of common tags increases to 18. We find five issues that two employees tagged differently, and we ask the employees what they think about these differences. After the exchange of ideas between each other, in two of the different tagged issues, the developer thinks that the tag value of the leader is more appropriate, and in the other two, the leader thinks that the tag value of the developer is more accurate. In an issue, they cannot reach a common decision. Finally, we add the values from the ITS to the combination and find the values where our model coincides with at least one of the labels of the two employees and the label from the ITS. Thus, we see that our model and ITS have the same label value with that undecided issue.

\begin{table}
\centering
 \caption{Comparison of label results.}
\begin{tabular}{|c|c|}
\hline
\textbf{Combination} & \textbf{\# labels}
\\\hline
    $\text{LabelByDeveloper} \cup \text{LabelByLeader} \cup \text{Model} \cup \text{ITS}$ & 11 
    \\
    $\text{LabelByDeveloper} \cup \text{LabelByLeader} \cup \text{Model}$ & 13 
    \\
    $\text{Model} \cup (\text{LabelByDeveloper} \lor \text{LabelByLeader})$ & 18
    \\
    $\text{Model} \cup (\text{LabelByDeveloper} \lor \text{LabelByLeader} \lor \text{ITS})$  & 19
\\\hline
\end{tabular}
\label{table:comparison}
\end{table}

In the last section, we direct the questions we prepared to the employees to get an idea about the system. We ask whether they would prefer such a system to be used in business life. They state that if they are converted into an application and the necessary features are added, for example, if an interface is provided where they can enter the current number of personnel, and their experiences, add and remove employees who are currently on leave, and if they turn into a plugin that integrates with the Jira interface, they will want to use it. Afterward, we ask if you find the system reliable and do you trust the assignments made. They say that they cannot completely leave the assignment to the application, and they will want to take a look at the assignments made by the application. The team leader adds that if there is a feature to send his approval, for example, by sending an e-mail before making the appointment, he will take a look and approve it, except for exceptional cases, and his work will be accelerated. As a result of the assignments made with the system, we address the question: Do you think that the average task solution time will decrease? It can reduce the average task resolution time, but they state that they think that if similar tasks are constantly sent to similar employee groups, this may have undesirable consequences for employee happiness and development. Next, we ask if you think using the system will reduce planning time. There are times when they talked at length in team planning meetings about who would get the job and who would be more suitable. At least, they think that if they have a second data, it can be a savior in cases where they are undecided. Finally, we would like to know your suggestions to improve the system. They state that if this system is going to turn into an application, they will want to see the values that the application pays attention to, to be able to edit and remove or add new ones. They think that if it has a comprehensive and user-friendly interface, it will still be suitable for use in business processes.

\section{Related Work}
\label{sec:relwork}
Several studies in the literature have focused on issue classification, which has addressed a variety of objectives, including issue assignment, effort estimation, issue prioritization, and so on. In this section, we briefly give details regarding issue assignment studies in general and all Turkish-language issue classification studies in particular.

Several types of research have been conducted in order to automate the time-consuming task of issue assignment. In 2017, Goyal et al. \cite{Goyal2017} review and categorize 75 research papers on the automated bug assignment area. They identify seven categories: machine learning \cite{Jonsson2016,Shokripour2015,Xia2017,Anvik2006,Bhattacharya2012,Oliveira2021}, information retrieval \cite{Yang2016}, auction \cite{Hosseini2012}, social network \cite{Yang2014}, tossing graph \cite{Su2021}, fuzzy set \cite{Panda2022} and operational research based \cite{Karim2016} techniques. They capture the fact that for automatic bug report assignment, machine learning and information retrieval techniques are the most popular ones. In recent years, deep learning algorithms have also been successfully applied in this field, which has recently revolutionized the idea of word sequence representation and demonstrated encouraging advancements in a number of classification tasks \cite{Goodfellow2016}. In this section, we restrict our focus to machine learning and deep learning architectures used to train issue assignment systems.

The machine learning algorithms use historical bug reports to build a supervised or unsupervised machine learning classifier, which is then used to choose appropriate developers for new bug reports. Naive Bayes is the most widely used classifier in machine learning-based approaches according to prior studies \cite{Xuan2010,Bhattacharya2012,Banitaan2013,Mahajan2022,Shokripour2015}, and it has been extensively tested \cite{Goyal2017} in the bug reports of open-source projects. Most studies use Eclipse \cite{Bhattacharya2012,Anvik2006,Aljarah2011,Sureka2012,Alenezi2013,Peng2017} and Mozilla \cite{Bhattacharya2012,Sureka2012,Peng2017,Hernandez2018} projects to validate their proposals. Machine learning models in most approaches \cite{Anvik2006,Oliveira2021,Sureka2012} use only summary and description as textual features of the issues. Jonsson et al. \cite{Jonsson2016} use the combined title and description as textual features and version, type, priority, and submitter columns as nominal features. Sharma et al. \cite{Sharma2017} consider bug attributes, namely, severity, priority, component, operating system, and the bug assignee. 

To estimate the value of terms, most of the approaches \cite{Sureka2012,Jonsson2016,Shokripour2015,Sajedi2020} in the literature employ term-weighting techniques like Tf-Idf. Jonsson et al. \cite{Jonsson2016} represent textual parts in the bug reports as the 100 words with the highest Tf-Idf. Shokripour et al. \cite{Shokripour2015} use time metadata in Tf-Idf (Time-Tf-Idf). To determine the value of terms in a document and corpus, the Tf-Idf technique only considers their frequency. However, in determining the weight, time-based Tf-Idf considers the time spent using the term in the project. The developer's recent use of the term is taken into account when determining the value of the developer's expertise. They rank the developers according to their calculated term expertise, and the first developer on the list is assigned to fix the new bug.

However, prior studies focused on open-source projects only but rarely \cite{Jonsson2016,Oliveira2021} attempted in industrial environments like our study. Jonsson et al. \cite{Jonsson2016} use ensemble learner Stacked Generalization, which is our best method also, that combines several machine learning classifiers on data from the automation and telecommunication company. In their approach, the different classes correspond to the development teams. Oliveira et al. \cite{Oliveira2021} also use the data set of a large electronic company. They create a model that can distribute new issues according to the responsibilities of the teams using a variety of machine learning techniques and the WEKA \cite{Weka2008} tool.

To improve prediction accuracy, some researchers use incremental learning methods. Bhattacharya et al. \cite{Bhattacharya2012} use various machine learning algorithms and achieve the best results using the NB classifier in combination with the product-component features, tossing graphs, and incremental learning in mostly used large projects: Mozilla and Eclipse. Xia et al. \cite{Xia2017} offer the multi-feature topic model (MTM), a specialized topic modeling approach that extends Latent Dirichlet Allocation (LDA) for the bug assignment. To map the term space to the subject space, their approach takes into account product and component information from issue reports. Then, they suggest an incremental learning mechanism that uses the topic distribution of a new bug report to assign an appropriate developer based on the reports that the developer has previously fixed. 

The deep learning algorithms are attempted first in 2017 \cite{Florea2017} for bug report assignment recommendation, to the best of our knowledge. Gupta et al. \cite{Gupta2021}  describe the popular deep learning approaches applied to the domain of bug reports and Recurrent Neural Networks (RNN) and Long Short Term Memory (LSTM) are a few famous approaches being used for the deep learning-based approaches \cite{Florea2017,Mani2019}. Mani et al. \cite{Mani2019} use title and description parts and Florea et al. \cite{Florea2017} use the component id, product id, and bug severity fields as one-hot-encoded categorical variables in addition to title, description, and comments to represent the issues. In 2022, Feng et al. \cite{Feng2022} use four transformers models BERT and RoBERTa along with their distilled counterparts DistilBERT, DistilRoBERTa in an ensemble using a resolver team, resolver person, and description columns of the issues.

In research using Turkish issue reports, there are limited studies available in a few fields. The reason may be the agglutinative nature of the Turkish language and the absence of a shared data set for Turkish issues in the literature. Aktas et al. \cite{Aktas2020} classified the issues they gathered from the banking industry among various software development teams. They use the Jira issue reports for their research like our study. They use SVC, CNN, and LSTM models to solve the classification problem, and they represent the summary and description columns of the issue reports as ordered vectors of Tf-Idf scores. The linear SVC classifier offers the best assignment accuracy for their research with a 0.82 score. Koksal et al. \cite{Koksal2021} present an automated bug classification approach using a commercial proprietary bug data set. They apply several machine learning algorithms and the SVM classifier is the best algorithm with 0.72 accuracy. In 2022, Tunali \cite{Tunali2022} prioritizes the software development demands of a private insurance company in Turkey. He proposes several deep-learning architectures and a pre-trained transformer model called distilbert-base-turkish-cased based on DistilBERT to achieve the highest accuracy of 0.82.

\section{Conclusions and Future Work}
\label{sec:conclusions}
This study focuses on automated issue assignment using proprietary issue reports obtained from the electronic product manufacturer's issue tracking system. The objective of the issue assignment approach is to assign issues to appropriate team members based on their respective fields. The team members are categorized into Software Developer, Software Tester, Team Leader, and UI/UX Designer. Among these categories, the majority of the data set consists of developers. Efficiently allocating issues to developers is critical for effective time management. To achieve this, we further classify developers into Senior, Mid, and Junior levels, which are widely accepted labels in the industry. 

Our focus lies in extracting features from the filled Jira columns, as well as the title and description texts of the issues, utilizing NLP techniques. These features serve as inputs to our learning methods, enabling us to analyze and classify the issues effectively. Additionally, we employ other commonly used word embedding methods which are Tf-Idf, BOW, and Word2Vec to generate feature vectors from the text fields. This step, implemented using the Sklearn and Gensim library, allows us to compare the performance of our feature set against alternative approaches. Furthermore, to assess the effectiveness of our overall methodology, we incorporate widely adopted deep-learning techniques, namely DistilBert, Roberta, and Electra. 

Following the production of feature vectors, we proceed to implement the proposed system utilizing established machine learning techniques. With the aim of enhancing predictive performance, we employ ensemble methods that leverage a diverse range of machine-learning algorithms. To evaluate the effectiveness of our system, we employ widely recognized metrics such as accuracy, precision, recall, and F1-score which serve as indicators of its performance. To further refine our predictions, we employ a robust technique known as 10-fold cross-validation. In order to conduct a thorough statistical analysis, we construct a matrix to compare and contrast the effectiveness of our proposed strategies. This matrix allows us to assess the performance of our system across different algorithms, ensemble techniques, and evaluation metrics. 

Our future endeavors involve the development of a versatile tool applicable to diverse software team models. To fortify our work, we actively engage in discussions and pursue collaborations to acquire data sets from businesses operating across various domains, such as game development and banking applications. This broadened data set will enable us to enhance our model's capabilities for multi-class classification, accommodating different roles within software teams, including product owners and business analysts. Furthermore, we are committed to ensuring compatibility and flexibility by incorporating various business branches into our data set. By incorporating real-world data obtained directly from industry sources, both in English and Turkish, we will conduct comprehensive evaluations through diverse studies. Expanding on the existing features, we intend to utilize the same data set for future research endeavors, such as effort estimation \cite{Weiss2007,Zhang2013}, further solidifying the value and applicability of our work in the field.

\bibliographystyle{splncs04}
\bibliography{paper}

\end{document}